\documentclass[runningheads]{llncs}
\usepackage[T1]{fontenc}
\usepackage{graphicx}
\usepackage[utf8]{inputenc}
\usepackage{amsmath}
\usepackage{booktabs}
\usepackage{algorithm}
\usepackage{algorithmic}
\usepackage{tabularx}
\usepackage{url}
\usepackage[hidelinks]{hyperref}
\begin{document}
\title{Gaussian Neural Networks}
\titlerunning{Gaussian Neural Networks}
%
\author{Peter Kuhn\inst{1} \and
Victoria Heusinger-He{\ss}\inst{1}}
\authorrunning{P. Kuhn, V. Heusinger-He{\ss}}
%
\institute{Fraunhofer Institute for High-Speed Dynamics, Ernst-Mach-Institut, EMI, Germany\\
\email{\{peter.kuhn,victoria.heusinger-hess\}@emi.fraunhofer.de}}
\maketitle              
\begingroup
\renewcommand{\thefootnote}{}
\footnotetext{An extended abstract for this paper has been accepted for publication at the International Conference on Neural Information Processing (ICONIP) 2026.}
\addtocounter{footnote}{-1}
\endgroup

\begin{abstract}
Gaussian neural networks (GaNNs) are proposed as a novel regularization mechanism for neural networks. From a Bayesian perspective standard regularization techniques can be viewed as imposing priors over weight-space. Assuming priors over activation-space remains a largely unexplored possibility. GaNNs assume such priors. They do this by treating activities from earlier layers like signals with Gaussian noise and predicting the properties of the noise distribution using an additional unsupervised loss. While training, the unsupervised loss acts as a penalty on unexpected activities, allowing greater weight updates in less surprising directions. The paper demonstrates the superiority of Gaussian neural networks over standard neural networks on a variety of classification and regression tasks. We also investigate the ability of GaNNs to quantify uncertainty.

\end{abstract}
\section{Introduction}

It is a surprising fact about standard neural networks that they cannot be surprised. They possess no explicit representation of what does and what does not constitute expected neuronal activation vectors. However, in order to learn effectively, it seems that prediction errors should drive less hypothesis revision where activities are expected versus where they come as a surprise. Some regularization techniques can be thought of as addressing this weakness implicitly, as they bias the network against learning over-complicated mappings, thereby disincentivizing networks to learn from (unsurprising) noise in the data~\cite{BNNs}. In a similar vein, neural networks by themselves are not well suited to deal with and quantify uncertainty inherent in the data (aleatoric uncertainty) or in model parameters (epistemic uncertainty). Here various techniques, from noise induction while learning~\cite{srivastava14a,gal2016} to ensemble methods~\cite{lakshminarayanan2017}, have proved useful.

This paper proposes a novel regularization technique that induces learnable priors in the activation space of a neural network. This is done by assuming that the activities of hidden layers themselves can be treated as noisy or uncertain. By amending the standard supervised learning objective by an additional unsupervised learning objective of the probability distribution over hidden node activities a model develops an intrinsic bias towards `normal' activity regimes. This results in a novel architecture, the Gaussian neural network (GaNN), that is intrinsically better regularized. Two ways to model noise of hidden units are proposed, resulting in low to medium computational overhead over standard architectures. Both architectures achieve better predictive performance across a variety of classification and regression tasks as compared to standard neural architectures trained with standard regularizers. By sampling from the learned activity-priors, the method can also be used to quantify predictive uncertainty, though the results here are mixed. GaNNs can be implemented with very low computational and memory costs. Depending on the details, they have only two additional parameters \textit{per neuron}. GaNNs alter some of the fundamental wiring of neural networks and can in principle be combined with any kind of neural wiring. In our experiments we intentionally restrict our attention to simple dense networks to validate the core architecture.

\section{Related Work}

\subsection{Bayesian Approaches to Neural Networks}

Early work on the deeper integration of Bayesian methods and neural networks took the form of Bayesian neural networks~\cite{BNNs}. These work by placing Gaussian prior distributions on the weights, tripling the number of parameters. The modeled uncertainty is the uncertainty of parameters or epistemic uncertainty. These networks can be trained using approximate Bayesian methods, chief among them variational inference. Bayesian neural networks inherently generate predictive distributions and can thus quantify the epistemic uncertainty of predictions, and are less prone to overfitting. However, these advantages are accompanied by a significant increase in necessary computing power. Also, the tripling of the parameter count can be debilitating in training very large models.

Dropout offers a way of approximating the functioning of Bayesian neural networks in standard architectures without increasing the number of model parameters. Dropout consists in the stochastic deactivation of nodes while learning. Intuitively, the induction of internal noise while training forces the network to learn a representation that is stabilized against that noise. This can be shown to approximate the functioning of Bayesian neural networks trained with variational inference~\cite{srivastava14a} and can be used to quantify predictive uncertainty using Monte-Carlo dropout, i.e.\ leaving dropout active while inference where the distribution over predictions corresponds to the predicted distribution~\cite{gal2016}. More recent work has emphasized differentiating between different kinds of uncertainty by modelling deeper properties of the target distribution instead of attempting deeper revisions of the model structure~\cite{PN,DER}. Such approaches are in principle compatible rather than in conflict with the approach proposed here.

\subsection{The Bayesian Perspective on Regularization}

If a learning algorithm is too flexible the learned mappings will generalize badly to unseen data. This simplicity bias has its parallel in ordinary life in Occam's razor. In neural learning, it typically shows up as regularization terms that are added to the loss function. Here, weight decay~\cite{decay} has been established as a strong default. The base loss $\mathcal{L}_{B}$, depending on the weights $\theta$ is amended by the L2-norm of the weights, connected by a hyper-parameter $\lambda$:

\begin{equation}
\mathcal{L}(\theta) = \mathcal{L}_{B}(\theta) + \frac{\lambda}{2}\,\|\theta\|_2^2
\label{eq:weight_decay}
\end{equation}

From the Bayesian perspective, weight decay can be understood as an isotropic Gaussian prior on weights such that lower weights have a higher prior probability~\cite{BNNs,deepbayes}.

\subsection{Gaussian Variance Heads}

An approach to uncertainty parallel to Bayesian neural networks and their approximate solutions (and indeed one that can be integrated with them~\cite{Kendall2017}) is to perform maximum-likelihood inference in the final model layer to model the noise inherent in the data, resulting in variance heads~\cite{nix1994}. This approach is standardly used to predict aleatoric uncertainty.

Variance heads predicting $\sigma^2(\vec{x}_i)$ are outputs of a model corresponding to data-point $\vec{x}_i$. It is assumed that the data points $d_i$ scatter around the model output $y(\vec{x}_i)$ as a Gaussian distribution parameterized by the weights $\theta$:

\begin{equation}
P(d_i|\vec{x}_i; \theta)=\frac{1}{\sqrt{2\pi \sigma^2(\vec{x}_i)}}e^{\frac{-(d_i-y(\vec{x}_i))^2}{2\sigma^2(\vec{x}_i)}}
\label{eq:gaussian_error}
\end{equation}

In a maximum-likelihood approach we would adjust network weights in order to maximize the probability of the data given the model, i.e.\ $P(d_i|\vec{x}_i; \theta)$. Because a function has its maxima where the logarithm of the function has its maxima, we can alternatively maximize the natural logarithm. Switching signs and taking the sample-mean we arrive at the Gaussian loss function (sometimes called Gaussian negative log-likelihood or Gaussian NLL):

\begin{equation}
\mathcal{L}_G = \sum_{i} \frac{1}{2}\left( \frac{(d_i - y(\vec{x}_i))^2}{\sigma^2(\vec{x}_i)} + \ln \left(\sigma^2(\vec{x}_i)\right)\right) 
\label{eq:gaussian_loss}
\end{equation}

Note that this is a more general form of the well known mean squared error and becomes equivalent to it for a uniform variance of $1$. Minimizing the full Gaussian loss a model will learn to predict both the mean $y(\vec{x}_i)$ as well as the variance $\sigma^2(\vec{x}_i)$ of the data-points $d_i$.

Note that the common assumption~\cite{Kendall2017} that this approach quantifies solely aleatoric uncertainty can be questioned. The total uncertainty of a prediction can be conceptualized as its expected error~\cite{epub92168}. Minimizing the expected error using maximum-likelihood inference, assuming it takes the form of a Gaussian, will be precisely equivalent to the described process of training a variance head. Learning the expected error of a model is an established method for measuring total uncertainty~\cite{lakshminarayanan2017,peter&daniel}. The only difference to a variance head is that it is typically integrated within one model rather than training a second higher-order model. The variance is trained based on the divergence of model output and prediction across trials and this divergence will necessarily result from both epistemic and aleatoric components.

\section{Gaussian Neural Networks}

\subsection{Inner Gaussian Loss}

We propose a new kind of neuronal architecture, the GaNN, that can be effectively trained using back-propagation and that has the promise to be intrinsically well-regularized with minimal computational overhead. GaNNs are regularized by isotropic Gaussian priors over activation-space. GaNNs consist of Gaussian hidden layers. Gaussian hidden layers operate under the assumption that the activity of the network can itself be treated like noisy signals. On a very high level this formalizes the idea that one should track the uncertainty of one's beliefs or representations. Going deeper, neuronal activities represent real-world features (even if determining what features exactly are represented can be very hard). The assumed internal noise then reflects the uncertainty about whether the neuronal activity accurately captures the represented feature. The GaNN will learn priors over activities, i.e.\ activity-priors, that can also be seen as priors regarding a represented feature. The violation of this prior will come with a penalty on the loss, entailing an imperative for the network to perform weight updates that conform to their learned prior probability.

Similar to Bayesian neural networks that represent weight and bias priors, pure Gaussian hidden layers possess Gaussian activity priors for all their neurons. Every layer $l$ consists of three vectors representing a mean-prior $\vec{\mu}^{(l)}$ and a variance-prior $\vec{\sigma}^{(l)^2}$, in addition to an activity $\vec{a}^{(l)}$. We here adopt the convention of using bracketed superscript indices to refer to either singular neurons or layers, which will be clear from context. The full state of a layer can be represented as a vector $\vec{s}^{(l)}=\left(\vec{\mu}^{(l)}, \vec{\sigma}^{(l)^2}, \vec{a}^{(l)} \right)^T$. To make the whole construction mathematically traceable we assume the noise at a layer to be independent of the noise at previous layers, resulting in the following likelihood for an $N$-layer model (the output layer is non-Gaussian), parameterized by the weights $\theta$:

\begin{equation}
\begin{aligned}
P(d_i|\vec{x}_i;\theta)=
P(d_i|\vec{s}^{(N-1)}_i;\theta)P(\vec{s}^{(N-1)}_i|\vec{s}^{(N-2)}_i;\theta)\cdots P(\vec{s}^{(1)}_i|\vec{x}_i;\theta)
\label{eq:independent_probabilities}
\end{aligned}
\end{equation}

Where $\vec{s}_i^{(k-1)}$ is the state corresponding to input sample $\vec{x}_i$. We assume the probability distribution for the activities for neuron $k$ to be Gaussian parameterized by that neuron's activity-prior. Just as in the case of variance heads, the variance-priors will be a learnable function of the previous layer's state, thus written as $\vec{\sigma}^{(k)^2}\left(\vec{s}_i^{(k-1)}\right)$. Thus, the neuron-wise likelihood will be Gaussian around the mean-prior with the respective variance:

\begin{equation}
P(s_i^{(k)}|s_i^{(k-1)}; \theta)=\frac{1}{\sqrt{2\pi \sigma^{(k)^2}\left( s_i^{(k-1)}\right)}}e^{\frac{-\left(\mu^{(k)}-a\left(s_i^{(k-1)}\right)\right)^2}{2\sigma^2\left(s_i^{(k-1)}\right)}}
\label{eq:gaussian_activity}
\end{equation}

We will assume that the input and output layer have no activity-priors and we will ignore them from now on. Taking the logarithm of Eq.~\ref{eq:independent_probabilities}, as the probabilities for activities in layers and neurons within layers are independent, with index $k$ looping over all neurons in all layers except the input- and output-layer, we get:

\begin{equation}
\begin{aligned}
\ln\left(P(d_i|\vec{x}_i;\theta)\right)= \ln P(d_i|\vec{s}^{(N-1)}_i;\theta) + \sum_{k}\ln\left(P(s_i^{(k)}|s_i^{(k-1)};\theta)\right)
\label{eq:new_gaussian_loss}
\end{aligned}
\end{equation}

Focusing on the right term, just as for the training of variance-heads we arrive at a loss function for training activity priors:

\begin{equation}
\mathcal{L}_I = \sum_{i,k} \frac{1}{2}\left( \frac{\left(\mu_i^{(k)} - a^{(k)}\left(\vec{s}_i^{(k-1)}\right)\right)^2}{\sigma^{(k)^2}\left(\vec{s}_i^{(k-1)}\right)} + \ln \left(\sigma^{(k)^2}\left(\vec{s}_i^{(k-1)}\right)\right)\right) 
\label{eq:unsupervised_loss}
\end{equation}

Learning Gaussian activity priors is a form of unsupervised learning as the Gaussian loss in Eq.~\ref{eq:unsupervised_loss} is independent of any data-points $d_i$. We will refer to it as the inner loss of the network. In order to facilitate real learning, the Gaussian loss needs to be amended with a data related base loss to get a semi-supervised total loss. Here, it is tempting to rely on the sum of unsupervised Gaussian loss and some base loss, however, this turns out to be inadvisable. Depending on the dataset, the Gaussian loss may be on a different order of magnitude compared to the base loss, leading the network to focus almost solely on learning activity priors. Fixed weights, due to differences in the evolution of the losses, are similarly inadvisable. Standard solutions for these kinds of multi-goal learning tasks involve additional learnable loss-weights \cite{DBLP:journals/corr/KendallGC17}. However, we note that it is easier to  define a no-gradient operator $\Theta$ such that, for any $g$ with $\theta \mapsto g(\theta)$:

\begin{equation}
\nabla_{\theta}\Theta\bigl(g(\theta)\bigr)=0
\label{eq:no_grad}
\end{equation}

This is implemented in PyTorch natively as the \texttt{no\_grad} operation. We can then define the loss of our network, using an arbitrary base loss $\mathcal{L}_{B}$, so that the relative contribution of the two loss functions to the total loss will only depend on a new hyper-parameter $\alpha$:

\begin{equation}
\mathcal{L}=\mathcal{L}_{B}+\alpha\Theta \left(\left|{\frac{ \mathcal{L}_B}{\mathcal{L}_{I}}}\right|\right)\mathcal{L}_I
\label{eq:loss}
\end{equation}

$\alpha$ specifies how much emphasis will be put on learning the target objective vs.\ learning activity-priors. As in Eq.~\ref{eq:weight_decay}, $\alpha$ also determines the strength of the regularization. As the inner loss can be negative, this may result in the counter-intuitive case of the total loss being zero across training. Considering the impact of $\Theta$, it should be clear that this will not negatively impact learning because there will still be non-zero derivatives owing to those parts of the equation outside the scope of $\Theta$.

The impact of the unsupervised loss is similar to the neuronal mechanism proposed to underlie human attention (see~\cite{friston}). The function of the inner loss can be conceptualized as a penalty on unexpected activities. The added network nodes learn what regime of activities is to be expected. The inner loss then quantifies and penalizes a gradient evolution outside the learned expected regime. In this way, expectations are integrated into the functioning of the network as larger steps are taken when the system is within an expected regime, while more cautious gradient updates are in order outside of it. As noted in the introduction, it should be surprising that standard neural networks do not learn any explicit representation of what constitutes normal activity. This would also result in a regularization effect as the activity priors disincentivize learning over-complicated mappings, relative to the mappings that have already been learned.

If these considerations are correct, we would expect a model that minimizes an inner loss to perform worse on the validation set in the beginning of training, as the model first has to learn reasonable activity-priors alongside minimizing the base loss. Once this initial learning was performed, the regularization effect should kick in and the network should converge more optimally as compared to a model merely learning the base-objective in a supervised fashion. We would not expect the inner loss to reduce indefinitely, as the learning of the base-objective will also shift the expected activities around, constantly shifting what the network is trying to learn. So the inner loss should reach a relatively stable plateau at some point in training or even increase where this can be canceled out by a corresponding minimization of the base loss.

\subsection{Dense and Sparse Gaussian Layers}

\begin{figure}[t]
  \centering
  \begin{minipage}{0.5\textwidth}
    \centering
    \includegraphics[width=\linewidth]{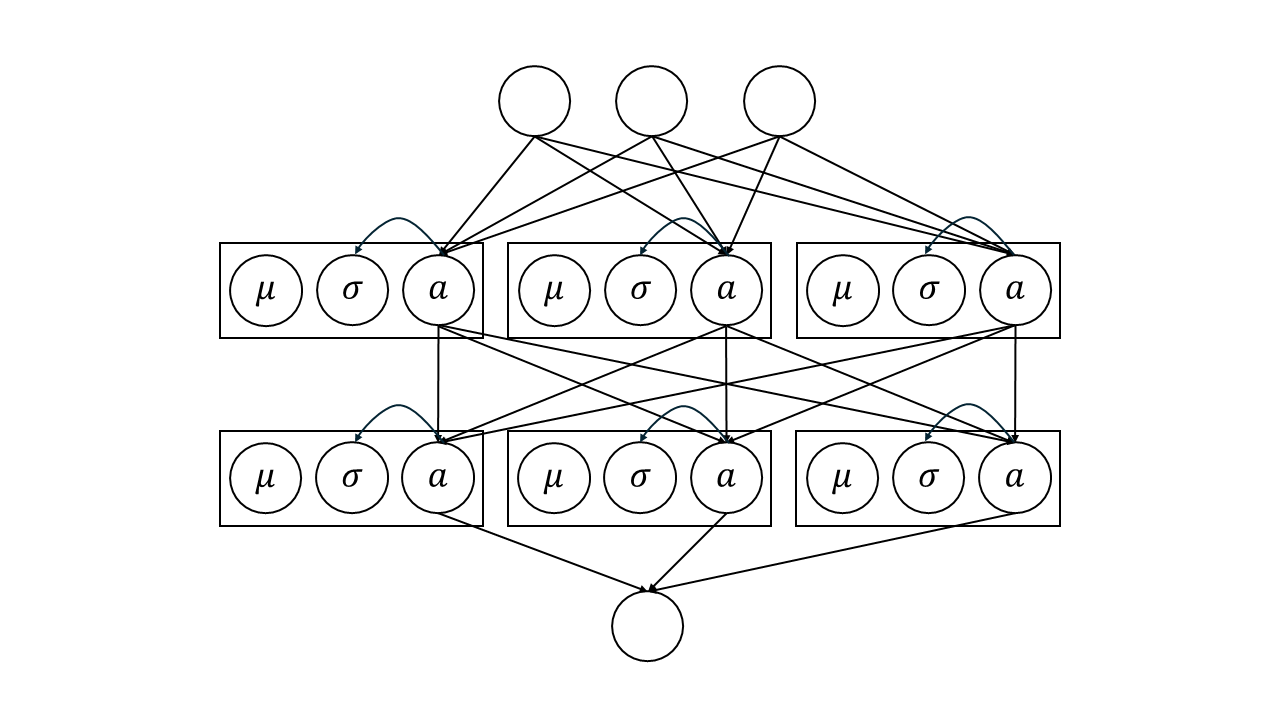}
    \par\vspace{2pt}(a)
  \end{minipage}\hfill
  \begin{minipage}{0.5\textwidth}
    \centering
    \includegraphics[width=\linewidth]{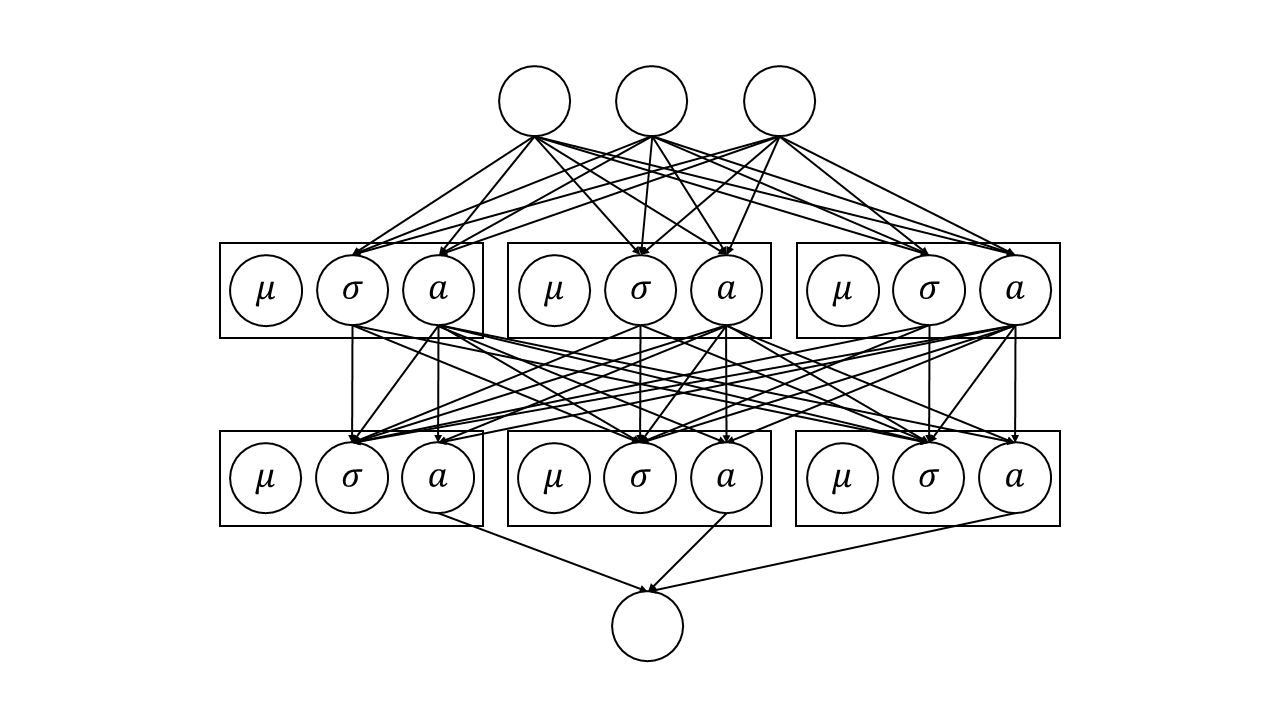}
    \par\vspace{2pt}(b)
  \end{minipage}
  \caption{Two models with two sparse (a) and dense (b) Gaussian hidden layers. Every box is a Gaussian neuron, containing an activity and an activity prior. In the sparse Gaussian layers, the variance-priors only depend on the associated activity. In the dense Gaussian layers, the variance-priors depend on all preceding nodes. In both architectures, the activities are independent of variances. The mean-priors only enter as constituents of the loss and are irrelevant in inference.}
  \label{fig_models}
\end{figure}

So far we have introduced an additional inner loss for learning Gaussian activity-priors. It would be possible to let the additional parameters be independent of the model input, similar to a bias in a normal network. However, if we want the network to learn more complex activity priors, we should make the variance terms functions of other activities. We will consider two such architectures, one sparse, resulting in minimal computational overhead, and one dense, which is able to learn more complex patterns but results in a medium to large computational overhead. (See Fig.~\ref{fig_models} for an overview.)

Variances are by definition positive and the inner loss becomes undefined otherwise. This can be ensured by using the right kind of activation function. Here, we will employ a softplus activation function $\zeta$ to ensure that:

\begin{equation}
\zeta(x) = \log\!\bigl(1 + e^{x}\bigr)
\label{eq:softplus}
\end{equation}

Furthermore, both the sparse and the dense model learn more stably using a hard-coded lower bound $v$ for the variance-prior as otherwise a variance approaching zero would entail an inner loss approaching infinity. Thus for the sparse model the variance-prior at layer $l$ will be (omitting dependencies on the input for readability):

\begin{equation}
    \vec{\sigma}^{(l)^2}_{sparse}=\zeta\left( \vec{a}^{(l)}\odot \vec{w}^{(l)}+\vec{b}^{(l)} \right) + v
\label{eq:sparse}
\end{equation}

$\vec{w}^{(l)}$ and $\vec{b}^{(l)}$ are weights and biases, of course not to be confused with the regular weights and biases mapping activities in one layer to activities in the next. $\odot$ represents the Hadamard product.

The dense noise model on the other hand works just as a normal dense network layer. Note however, that the activity in a layer does \textit{not} depend on the variance-priors of the previous layer in the same way. Forward-propagation for activities is defined in the usual way.

\begin{equation}
    \vec{\sigma}^{(l)^2}_{dense}=\zeta\left( W^{(l)}\,\vec{a}^{(l-1)} + K^{(l)}\,\vec{\sigma}^{(l-1)} +\vec{b}^{(l)} \right) + v
\label{eq:dense}
\end{equation}

$W^{(l)}$ and $K^{(l)}$ are again additional weight matrices.

The dense model adds considerable computational complexity and memory requirements, but can also learn more complex activity priors. The number of parameters grows quadratically with the number of Gaussian nodes. On the other hand, in case of the sparse architecture, the number of additional parameters when compared to a normal neural network is just three times the number of nodes, adding a mean-prior $\mu^{(k)}$, one connection weight as part of $\vec{w}^{(l)}$ and one bias term in $\vec{b}^{(l)}$. The computational and memory overhead will thus shrink the larger the model gets, becoming vanishingly small for very large models. The computational overhead of the sparse architecture is considerably lower than for ensemble techniques or Bayesian neural networks.

All added complexity of a GaNN as opposed to a standard neural network are only relevant for training and uncertainty quantification. Once training is done, all non-standard parameters can be dropped and the remaining units can be used for feed-forward inference.

We would predict that, overall, a GaNN equipped with a dense noise model will generate more powerful effects as opposed to the sparse noise model for it can encode more complex expectations about what constitutes ordinary activities, and what constitutes a surprise and should be penalized. On the other hand we may also expect that a too powerful noise model might focus on minimizing the inner loss indefinitely delaying progress on the main learning objective, causing slow convergence or even no convergence within the relevant time frames.

\subsection{Uncertainty Quantification}

\subsubsection{Noisy Sampling.}

A trained GaNN can perform uncertainty quantification by producing a distribution of predictions. This is similar, though not equivalent, to the process of drawing predictive samples in Monte Carlo dropout. As a GaNN treats its own activities like a noisy data-source, we can straight away sample from the learned noise distributions and add those to the relevant activities in inference. So, $\vec{s}_{\gamma,j}^{(l)}\left(\vec{x}_i\right)$ is the $j$th noisy sample of the $l$th state vector for the $i$th input sample. As the state vectors at layer $l$ only depend on the state vector at layer $l-1$ we say that:

\begin{equation}
    \vec{a}_{\gamma,j}^{(l)}= \vec{a}_{j}^{(l)}\left(\vec{a}_{\gamma,j}^{(l-1)}\right) + \gamma\varepsilon \left(0, \vec{\sigma}^{(l)^2}\left(\vec{a}_{\gamma,j}^{(l-1)}\right)\right)
\label{eq:noisy_sampling}
\end{equation}

$\varepsilon(\mu, \sigma^2)$ is a Gaussian noise term with mean $\mu$ and variance $\sigma^2$. For the input layer there will be no noisy sampling of course. $\gamma$ is a hyper-parameter that controls the strength of the noise. While the resulting distribution will naturally reflect predictive uncertainty, there is no reason to assume it should be well calibrated by default. We note that, in principle, $\gamma$ can be learned from data, though we do not attempt this here.

One might expect, generalizing from the example of variance heads, that the process will primarily help to quantify aleatoric uncertainty. However, as argued above, whether variance heads really solely quantify aleatoric uncertainty can be questioned. Another way to see this is to note that, if a network predicts high activity variance for some input, the relevant input will be unusual. Thus, if a data-point is distant from the training data, and thus the connected predictions are \textit{epistemically} uncertain, this should lead to high predicted variance.

\subsubsection{Performance Metrics.}

Uncertainty quantification suffers from the fact that the ground truth, except in the case of simulated data, is unavailable. We thus have to specify additional performance metrics to measure the quality of uncertainty quantification. In case of regression, we can simply calculate the mean and variance of the predictive distribution generated by drawing samples from the process specified by Eq.~\ref{eq:noisy_sampling}. We can then rely on the Gaussian negative log-likelihood or Gaussian NLL, already specified in Eq.~\ref{eq:gaussian_loss}, measuring the error on the test set with respect to the predicted mean and variance.

In case of classification we will start out by calculating the predicted probabilities $p_{k,i}$ for all classes $k$ and all samples $i\in[0,n]$ by taking the mean across the softmax output across predictive samples $j$. We will define the predicted class $\hat{d}_i$ as that with the maximum probability. These we can compare to the true classes $d_{i}$ in two ways that will tell us about how well the model captures uncertainty.

Just as in the case of the regression task, we can use the negative log-likelihood of the results, given the predicted probabilities as a guide to how well the ground truth (the true class $y_i$) was captured:

\begin{equation}
\mathrm{NLL} = -\frac{1}{n} \sum_{i=1}^n \log p_{y_i,i}
\label{eq:nll}
\end{equation}

Also, we may want to know how well we can detect whether our model is going to make an error from how uncertain it is. The relevant uncertainty per sample $i$ is equal to the entropy:

\begin{equation}
H_i = -\sum_{k} p_{k,i}\,\log p_{k,i}
\end{equation}

Then to assess how good a guide the uncertainty or entropy is to the probability of error we evaluate the probability that a randomly chosen error has a higher uncertainty score than a randomly chosen correct example (with ties counted as half). This results in the $\mathrm{AUROC}$ score (the `Area Under the Receiver Operating Characteristic curve') for error detection:

\begin{equation}
\mathrm{AUROC} =
\frac{1}{|E|\,|C|} \sum_{i \in E} \sum_{j \in C}
\Big( \mathbf{1}(H_i > H_j) + \tfrac{1}{2}\mathbf{1}(H_i = H_j) \Big)
\label{eq:auroc_err}
\end{equation}

$\mathbf{1}(P)$ is the fraction of samples for which condition $P$ holds. $C$ is the set of correct samples, $E$ is the set of errors, thus $|E|+|C|=n$.

Differentiating between different kinds of uncertainty is not the subject of this paper and the proposed metrics allow no such disentanglement.

\section{Experiments}

\subsection{Model Implementation}

All models were implemented in PyTorch. For the dense noise model, a layer can be implemented using PyTorch's standard Linear layer with a doubled number of units. After a forward pass through one layer the units are then split in half with one half representing activities, the other representing variances. Both are then fed into different loss functions, ReLU in the case of the activities, softplus in the case of variances. The sparse architecture is implemented with the help of a custom layer object. All Gaussian layers return their internal state vector $\vec{s}$, a list of which is added to the network's output for the computation of the supervised loss. The classical neural networks used for comparison are identical to the GaNNs, except for the additional structure. All models possess three hidden layers containing 1024, 512 and 256 units respectively; if dropout is used, there is one dropout layer per hidden layer. For Gaussian networks the dropout layers only attach to the activity nodes of the next layer, not to the variance nodes.

\subsection{Training Setup}

All experiments were performed five times with different random seeds and accumulated as a mean. Training was performed using stochastic gradient descent. A very simple and uniform learning rate schedule was used. For regression tasks, the initial learning rate was chosen to be $0.1$, for classification tasks $0.01$. After $30$ epochs it was divided by $10$. In some cases, the training was not stable (for both the standard and the Gaussian architecture) and gradient normalization to a norm of $10$ was performed to produce a training setup that could be applied universally to all models. The base loss for regression is MSE, the base loss for classification is cross-entropy. As with any regularization term added to the base loss, the strength of regularization, here encoded by $\alpha$, is a sensitive hyper-parameter. While $\alpha = 1$ can be a good default on many datasets, we find that this can place too strong constraints on learning. In case of simple classification tasks we thus used $\alpha=0.1$, in case of CIFAR100 and Tiny ImageNet we chose $\alpha=0.01$. The variance lower bound we chose to be $v = 0.5$ for all experiments. Where weight decay was used, we used a factor of $\lambda = 0.0001$. The chosen dropout rate was 0.3 for classification and 0.1 for regression tasks. The numbers of epochs used for specific datasets are listed in Table~\ref{tab:dataset_summary}. For test set evaluation we used the weight configuration where the model achieved maximal validation performance.

\subsection{Datasets}

We use ten datasets, five involving a regression, five involving a classification task. An overview is shown in Table~\ref{tab:dataset_summary}. Datasets are both visual and tabular to demonstrate the independence of performance of any specific data type. Regression datasets include the airfoil self-noise dataset created by NASA for the acoustic properties of airfoil blade sections~\cite{airfoil}, the yacht dataset where the hydrodynamic properties of yachts are predicted from dimensions and velocity~\cite{yacht}, the UTK face dataset for the prediction of ages from portrait pictures consisting of 64x64 color images~\cite{utkface}, the wine quality dataset~\cite{wine} where features are chemical properties of wine and the target is a taste score, and finally the million songs dataset where the year of publication is to be predicted from other features of the song. On the classification side, the CIFAR100 is composed of 32x32 color images that belong to 100 different classes~\cite{cifar} and Tiny ImageNet is composed of 64x64 color images that belong to 200 classes, the MNIST datasets are 28x28 greyscale images of handwritten digits~\cite{digits} and ten types of fashion articles~\cite{fashion}.

The MNIST datasets, Tiny ImageNet and CIFAR100 come with a predefined train-test-validation split. All other datasets were split into training and validation data (with different random seeds for different trials, see below) using an 60/20/20-split. 

\begin{table}[t]
\centering
\caption{Summary of dataset properties and number of epochs employed.}
\label{tab:dataset_summary}
\begin{tabularx}{\textwidth}{lXccc}
    \toprule
    \textbf{Name} & \textbf{Domain} & \textbf{N} & \textbf{Features} & \textbf{Epochs} \\
    \midrule
    \multicolumn{5}{l}{\textbf{Regression}} \\
    \midrule
    Airfoil & tabular & 1503 & 5 & 3000 \\
    Yacht & tabular & 308 & 6 & 5000 \\
    UTKFace & vision & 23708 & 12288 & 120 \\
    Wine Regression & tabular & 6497 & 11 & 120 \\
    YearPredictionMSD & tabular & 515345 & 90 & 120 \\
    \midrule
    \multicolumn{5}{l}{\textbf{Classification}} \\
    \midrule
    CIFAR-100 & vision & 60000 & 3072 & 120 \\
    Digits-MNIST & vision & 1797 & 64 & 120 \\
    Fashion-MNIST & vision & 70000 & 784 & 120 \\
    Wine Classification & tabular & 6497 & 11 & 120 \\
    Tiny ImageNet & vision & 100000 & 12288 & 120 \\
    \bottomrule
\end{tabularx}
\end{table}

\section{Results}

\subsection{Overall Performance}

\begin{figure}[t]
\centering
\includegraphics[width=\textwidth]{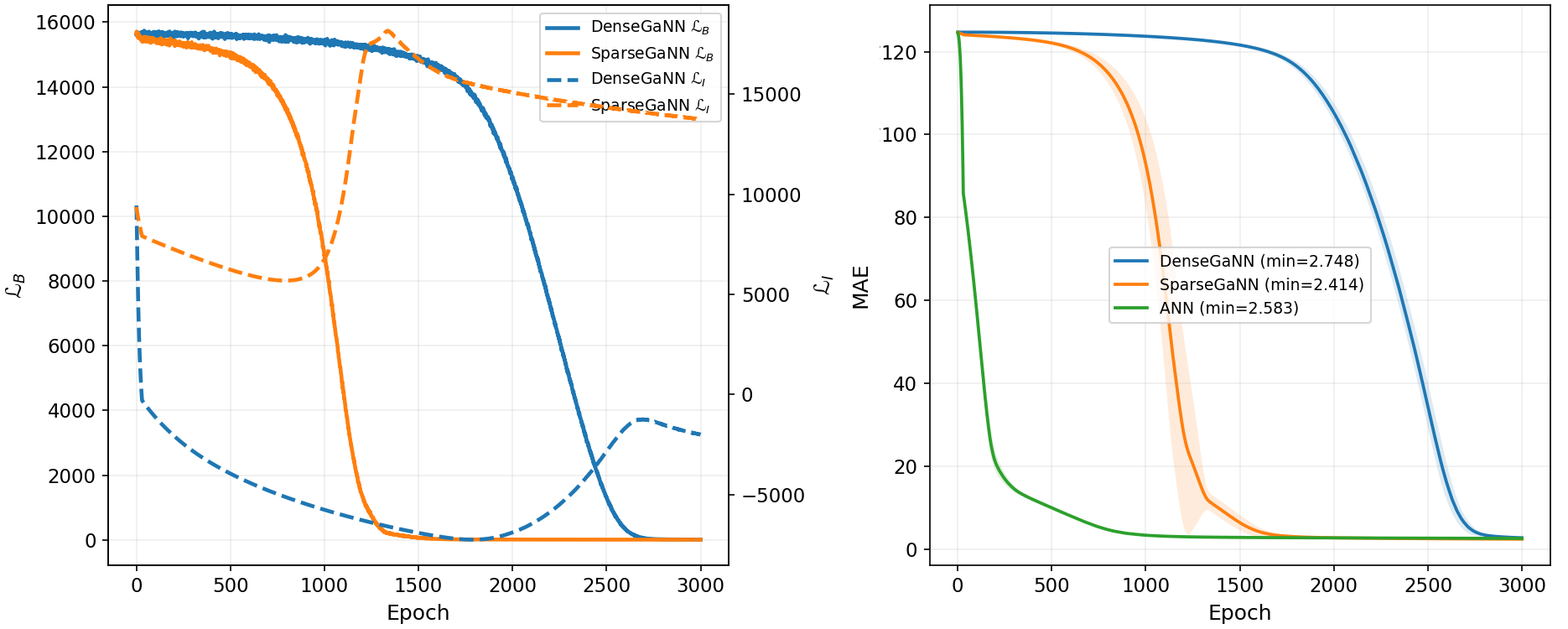}
\caption{Performance on the airfoil dataset, standard deviations are visualized as shaded regions. This dataset constitutes an extreme example of the predicted delayed convergence resulting in higher final performance. Delayed convergence is caused by first optimizing inner loss (left, dashed) and then optimizing the base loss.}
\label{fig:airfoil}
\end{figure}

Quantitative predictive performance results are summarized in Table~\ref{tab:perf_combined}. We observe first of all that GaNNs outperform the basic ANN consistently. The sparse architecture outperforms the ANN on nine out of ten datasets with one tie. The dense architecture does so in seven with one tie. Merely looking at overall performance across all experiments, this makes the performance of GaNNs more consistent than either weight decay or dropout. We also observe that the evolution of the loss follows our predicted trend: The networks first learn slower than rival networks, minimizing the inner loss in addition to the base loss. At some point the inner loss plateaus and the model focuses on minimizing the base loss. The inclusion of the learned knowledge about what constitutes normal activity enables higher peak performances as compared to non-Gaussian architectures. This trend is especially clear in the Airfoil dataset, visualized in Fig.~\ref{fig:airfoil}.

\begin{table}[t]
    \centering
    \caption{Test performance (mean for best model $\pm$ std). Top: Regression (MAE). Bottom: Classification (Accuracy). Note that for regression datasets, the ANN dropout columns use the variance-head variants.}
    \label{tab:perf_combined}
    \resizebox{\textwidth}{!}{%
    \begin{tabular}{lccccccc}
        \toprule
        \textbf{Dataset} & \textbf{GaNN-dense} & \textbf{GaNN-sparse} & \textbf{GaNN-sparse + D} & \textbf{ANN} & \textbf{ANN + D} & \textbf{ANN + WD} & \textbf{ANN + WD + D} \\ 
        \midrule
        \multicolumn{8}{l}{\textbf{Regression (MAE)}} \\
        Airfoil & $2.706 \pm 0.099$ & $2.335 \pm 0.075$ & $2.853 \pm 0.075$ & $2.494 \pm 0.090$ & $11.516 \pm 0.416$ & $2.494 \pm 0.090$ & $11.515 \pm 0.416$ \\
        Yacht & $2.208 \pm 0.154$ & $0.804 \pm 0.119$ & $1.173 \pm 0.107$ & $0.879 \pm 0.119$ & $5.521 \pm 0.386$ & $0.880 \pm 0.120$ & $5.521 \pm 0.386$ \\
        UTKFace & $10.164 \pm 0.094$ & $10.131 \pm 0.124$ & $10.227 \pm 0.116$ & $10.183 \pm 0.078$ & $16.719 \pm 0.168$ & $10.184 \pm 0.078$ & $16.719 \pm 0.168$ \\
        Wine Regression & $0.573 \pm 0.008$ & $0.589 \pm 0.008$ & $0.596 \pm 0.009$ & $0.613 \pm 0.005$ & $0.929 \pm 0.022$ & $0.613 \pm 0.005$ & $0.929 \pm 0.022$ \\
        YearPredictionMSD & $6.173 \pm 0.025$ & $6.159 \pm 0.025$ & $6.841 \pm 0.217$ & $7.029 \pm 0.062$ & $33.727 \pm 1.161$ & $7.024 \pm 0.061$ & $33.075 \pm 1.157$ \\
        \midrule
        \multicolumn{8}{l}{\textbf{Classification (Accuracy)}} \\
        CIFAR100 & $0.276 \pm 0.003$ & $0.278 \pm 0.003$ & $0.269 \pm 0.003$ & $0.276 \pm 0.004$ & $0.268 \pm 0.001$ & $0.278 \pm 0.002$ & $0.267 \pm 0.001$ \\
        MNIST-digits & $0.981 \pm 0.001$ & $0.980 \pm 0.001$ & $0.984 \pm 0.001$ & $0.980 \pm 0.001$ & $0.984 \pm 0.001$ & $0.980 \pm 0.001$ & $0.984 \pm 0.001$ \\
        MNIST-fashion & $0.900 \pm 0.001$ & $0.898 \pm 0.001$ & $0.899 \pm 0.002$ & $0.897 \pm 0.002$ & $0.899 \pm 0.001$ & $0.898 \pm 0.001$ & $0.899 \pm 0.001$ \\
        Wine Classification & $0.583 \pm 0.011$ & $0.576 \pm 0.013$ & $0.564 \pm 0.018$ & $0.574 \pm 0.013$ & $0.563 \pm 0.018$ & $0.570 \pm 0.014$ & $0.564 \pm 0.018$ \\
        Tiny ImageNet & $0.121 \pm 0.003$ & $0.119 \pm 0.001$ & $0.127 \pm 0.001$ & $0.118 \pm 0.002$ & $0.127 \pm 0.003$ & $0.122 \pm 0.001$ & $0.129 \pm 0.003$ \\
        \bottomrule
    \end{tabular}}
\end{table}

We observe that the dense noise model is not necessarily superior to the sparse one. In fact, the sparse model more often outperforms the dense variant than vice versa. Also, we observe that the dense model sometimes converges very late in training, for instance in case of the Airfoil and Yacht datasets. Both observations can be explained by an overpowerful noise model: It turns out to primarily minimize the inner loss for long stretches of training (see Fig.~\ref{fig:airfoil}). When convergence is achieved the model typically achieves better performances than any of the comparison models. We also observe that sparse GaNNs can sometimes be combined with dropout in a beneficial way, underlining that these are not strictly rival techniques.


\begin{figure}[t]
\centering
\includegraphics[width=\textwidth]{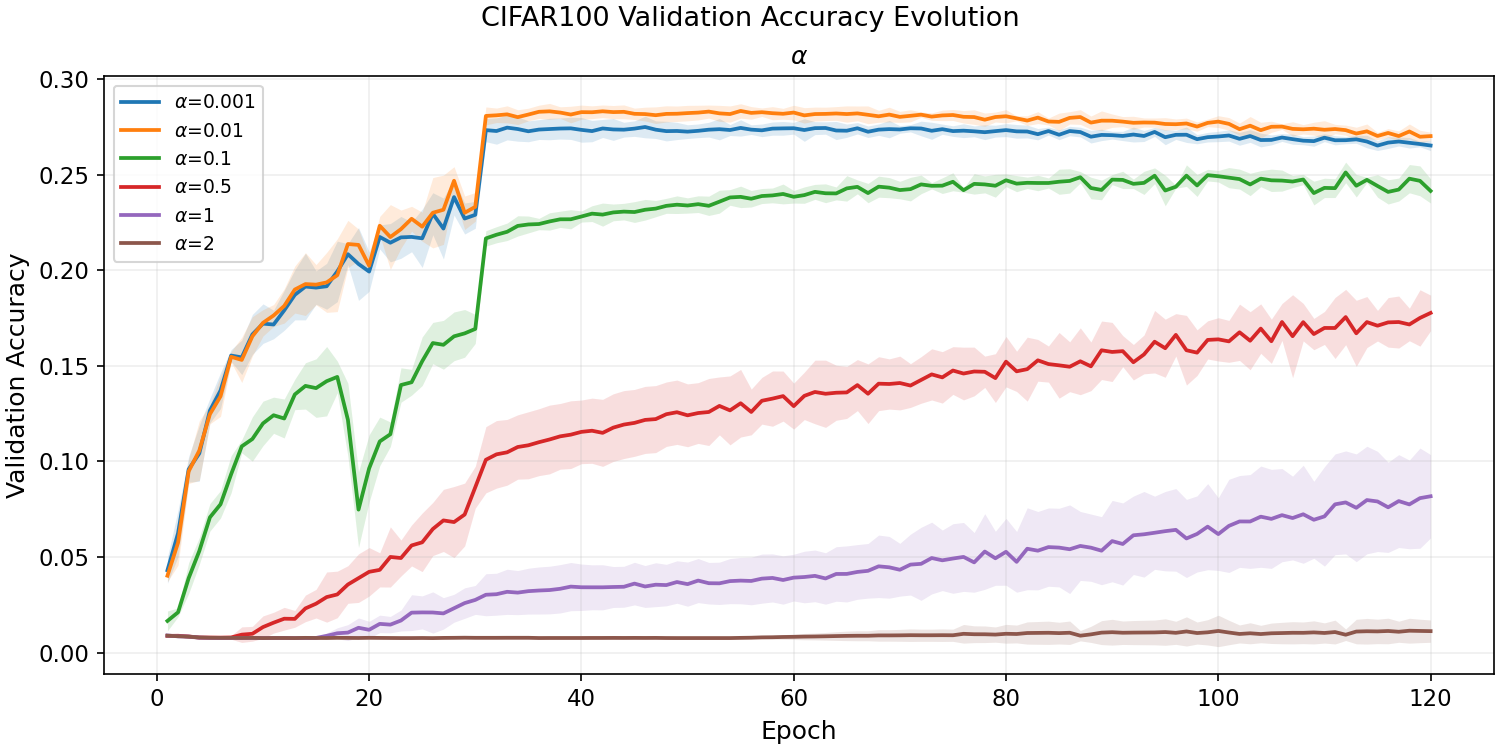}
\caption{Ablation studies on CIFAR100 using a sparse GaNN. Validation accuracy evolution for different values of $\alpha$.}
\label{fig:ablation}
\end{figure}

GaNNs introduce two novel hyperparameters, the loss-weight $\alpha$ and the variance lower bound $v$. A sweep for $\alpha$ on CIFAR100 (figure \ref{fig:ablation}), as well as our general experience while training,  indicates that $\alpha$ is easy to tune. Only high values ($0.5$ and above) cause a collapse in performance. Performance turns out, within a reasonable range, to be insensitive to $v$ --- its main function is to ensure numerical stability.

\subsection{Uncertainty Quantification}

GaNNs can be used for uncertainty quantification, as can be gleaned from the results shown in Table~\ref{tab:uq}. In case of regression, both GaNNs outperform their competitors. The best overall performance is arguably achieved by a combination of GaNNs with dropout. In the case of classification, the results are far less consistent.

\begin{table}[t]
    \centering
    \caption{Test UQ (means). Top: Regression (Gaussian NLL). Bottom: Classification (NLL / AUROC).}
    \label{tab:uq}
    \resizebox{\textwidth}{!}{%
    \begin{tabular}{lccccc}
        \toprule
        \textbf{Dataset} & \textbf{GaNN-dense} & \textbf{GaNN-sparse} & \textbf{GaNN-sparse + D} & \textbf{ANN + varhead} & \textbf{ANN + varhead + D} \\
        \midrule
        \multicolumn{6}{l}{\textbf{Regression (Gaussian NLL)}} \\
        Airfoil & 3.36 & 3.28 & 3.26 & 4.05 & 4.10 \\
        Yacht & 2.79 & 1.99 & 2.01 & 3.02 & 3.06 \\
        UTKFace & 12.92 & 7.57 & 7.89 & 4.50 & 4.50 \\
        Wine Regression & 1.23 & 1.27 & 1.28 & 1.61 & 1.63 \\
        YearPredictionMSD & 5.05 & 5.05 & 5.18 & 5.38 & 6.06 \\
        \midrule
        \multicolumn{6}{l}{\textbf{Classification (NLL / AUROC)}} \\
        CIFAR-100 & 3.17 / 0.750 & 3.17 / 0.746 & 3.02 / 0.734 & 3.20 / 0.751 & 3.04 / 0.735 \\
        Digits-MNIST & 0.11 / 0.954 & 0.12 / 0.953 & 0.13 / 0.953 & 0.08 / 0.974 & 0.06 / 0.975 \\
        Fashion-MNIST & 0.36 / 0.853 & 0.44 / 0.828 & 0.47 / 0.829 & 0.34 / 0.896 & 0.30 / 0.903 \\
        Wine Classification & 1.02 / 0.592 & 1.01 / 0.599 & 1.03 / 0.577 & 1.02 / 0.595 & 1.03 / 0.566 \\
        Tiny ImageNet & 4.77 / 0.702 & 4.81 / 0.706 & 4.18 / 0.706 & 4.93 / 0.709 & 4.19 / 0.705 \\
        \bottomrule
    \end{tabular}}
\end{table}

\section{Discussion}

The consistent positive performance impact of the GaNN architecture makes them both theoretically and practically interesting. In our experiments, the sparse noise model proved superior to the dense one, not just because of computational and memory efficiency, but because of a tendency towards more stable convergence.

The generally better regression results might speak to the fact that the regression models fit our Gaussian assumptions better. On the other hand, this paper focused on the usage of \textit{pure} Gaussian models. In practice, it might make sense to combine standard architectures with a few Gaussian layers to reap all positive effects. Also, the investigation of deeper models as well as the interaction of internal Gaussian models with convolutions and attention mechanisms might prove useful fields of further study. Finally, we note that the specifically Gaussian assumptions may be generalized to other kinds of distributions.

\begin{credits}
\subsubsection{\discintname}
The authors have no competing interests to declare that are relevant to the content of this article.
\end{credits}

\end{document}